%% file: arxiv.tex
\documentclass[letterpaper]{article}
\usepackage[preprint]{aaai2027}
\usepackage[hyphens]{url}
\usepackage{graphicx}
\usepackage{natbib}
\usepackage{caption}
\usepackage{algorithm}
\usepackage{algorithmic}
\usepackage{newfloat}
\usepackage{listings}
\usepackage[most]{tcolorbox}
\usepackage{booktabs}
\usepackage{xcolor}
\usepackage{soul}
\usepackage{multirow}
\usepackage{amsfonts}

\DeclareCaptionStyle{ruled}{labelfont=normalfont,labelsep=colon,strut=off} %
\floatstyle{ruled}
\newfloat{listing}{tb}{lst}{}
\floatname{listing}{Listing}

\newtcblisting{promptbox}[1]{%
    breakable,%
    listing only,%
    title={#1},%
    colback=black!5!white,%
    colframe=black!75!white,%
    colbacktitle=black!10!white,%
    coltitle=black,%
    fonttitle=\bfseries,%
    boxrule=0.5pt,%
    arc=0pt,%
    left=1mm,right=1mm,top=1mm,bottom=1mm,%
    before skip=0.75em,after skip=0.75em,%
    listing options={%
        basicstyle=\footnotesize\ttfamily,%
        numbers=none,%
        xleftmargin=0pt,%
        breaklines=true,%
        breakindent=0pt,%
        breakatwhitespace=false,%
        columns=fullflexible,%
        keepspaces=true,%
        aboveskip=0pt,%
        belowskip=0pt,%
        showstringspaces=false,%
        tabsize=2}}

\title{VideoLVR: Latent Visual Reasoning over Grounded Video Evidence}
\author{
    Shilong Li\equalcontrib,
    Xingyuan Bu\equalcontrib\corresponding,
    Hao Shi\equalcontrib,
    Yun Xing,\\
    Weixin Feng,
    Qiangpeng Yang,
    Shilei Wen
}
\affiliations{
    Beijing University of Posts and Telecommunications\\
    Tsinghua University\\
    Nanyang Technological University
}

\newif\ifrlshowchanges
\rlshowchangesfalse

\begin{document}
\maketitle

\input{sections/0_abstract_add_rl}
\input{sections/1_intro_add_rl}
\input{sections/2_relatedwork_add_rl}
\input{sections/3_methods_add_rl}
\input{sections/4_exp_add_rl}

\input{sections/5_conclusion_add_rl}

\bibliography{aaai2027}

\clearpage
\appendix

\section{Supervised Training Data Statistics}

The supervised training set contains 8,782 evidence-grounded reasoning
trajectories. Each trajectory is interleaved with one or more spatio-temporal
visual evidence groups, with each group associating an intermediate reasoning
claim with one or more localized visual regions. In total, the supervised training set
contains 25,710 evidence groups and 32,180 localized visual regions,
corresponding to 2.93 groups and 3.66 regions per trajectory, respectively.

\begin{table}[h]
\centering
\begin{tabular}{@{}p{0.73\columnwidth}r@{}}
\toprule
\textbf{Statistic} & \textbf{Value} \\
\midrule
Training Trajectories & 8,782 \\
Avg. Groups & 2.93 \\
Avg. Regions & 3.66 \\
Multi-Region Groups & 19.8\% \\
Multi-Moment Groups & 14.0\% \\
\bottomrule
\end{tabular}
\caption{Statistics of the supervised training set. Both averages are computed
over all evidence-grounded reasoning trajectories; the last two rows report
percentages over all evidence groups.}
\label{tab:training_data_statistics}
\end{table}

As shown in Table~\ref{tab:training_data_statistics}, each trajectory contains
2.93 evidence groups and 3.66 localized visual regions on average. The larger
number of regions reflects that some evidence groups are supported by multiple
localized visual regions. In particular, 19.8\% of the groups contain multiple
regions, and 14.0\% span multiple video moments. The training data therefore
includes groups that capture spatially distributed evidence within a frame or
temporally distributed evidence across frames, providing supervision for the
grouped latent visual evidence used by VideoLVR.

\section{Prompts for Evidence-Grounded Trajectory Synthesis}

We use two prompts in the data-synthesis pipeline described in
Section~\ref{sec:data_construction}. The first directly generates a reasoning
trajectory interleaved with spatio-temporal visual evidence groups, while the
second polishes a rejection-sampled trajectory by relocating each evidence
group to the reasoning step that uses it.

\subsection{Evidence-Grounded Trajectory Rejection Sampling}

\begin{promptbox}{System Prompt}
You are a video understanding expert. Your task is to watch the video carefully and answer the question. When answering, reason step by step according to the requirements and rely as much as possible on visible visual evidence.

Mark key visual evidence using:

<evidence claim="a short judgment supported by this group of visual evidence">
<box t="timestamp" content="a brief description of the bbox region">[x1,y1,x2,y2]</box>
...
</evidence>

Each evidence group supports one short, specific claim and may contain one or more boxes from the same or different video moments. Insert the group directly into the reasoning sentence whose judgment it supports, rather than presenting a separate evidence list. Use at most 10 groups and prefer evidence most relevant to the answer.

Inspect the provided video directly. Do not invent visual facts, timestamps, or coordinates. Copy each timestamp exactly from a visible video-context timestamp marker. Use tight, valid bounding boxes that exclude irrelevant background and do not cover almost the entire frame. Output normalized coordinates in [0,1], with the origin at the top-left.

For a multiple-choice question, place the reasoning inside <think>...</think>, followed by exactly one option letter and no additional text. Follow the required format strictly.
\end{promptbox}

\begin{promptbox}{User Prompt}
Video question:
{question}

Options:
{options}

Please follow the system instructions, reason carefully, and output the final answer as exactly one option letter.
\end{promptbox}

\subsection{Evidence-Guided Trajectory Polishing}

\begin{promptbox}{System Prompt}
You are a professional reasoning-data synthesis expert. Think privately and deeply, but expose only the requested JSON object in your final response.

Task
Reorganize the rejection-sampled trajectory so that each evidence group appears immediately before the reasoning statement it supports. Preserve the original reasoning content, rewriting surrounding text only as needed to integrate the relocated group. The result should read like a capable model naturally solving the question from the visual evidence, not like an edit report.

Reasoning-quality requirements
- Preserve every useful observation, comparison, option elimination, temporal/spatial relation, and inference in the original trajectory.
- Preserve meaningful false starts, provisional hypotheses, uncertainty, repeated verification, and self-correction. Do not turn tentative statements into certainty.
- Preserve the original reasoning order and wording whenever possible. Relocate each evidence group to the first reasoning step that uses its visual content.
- Preserve the original factual and logical content and final answer. Do not omit substantive reasoning or invent visual facts.
- Start directly with solving the problem. Do not mention data construction, prompts, placeholders, JSON, evidence IDs, model limitations, or the mechanics of selecting timestamps and boxes.

Protected evidence placeholders
- Tokens such as {{E001}} represent exact evidence XML that Python restores later.
- Use every supplied placeholder exactly once, character-for-character. Never invent one.
- Place each placeholder where its visual content is first inspected, before the resulting claim or conclusion. Do not collect placeholders at the end. All placeholders must precede the final answer synthesis.
- Use the Evidence Catalog to understand what each placeholder shows, but never copy coordinates, timestamps, IDs, XML, or catalog formatting into prose.

Output contract
- Return one valid JSON object with exactly two keys: think_content and answer.
- think_content is the complete reasoning body only. It must not contain <think>, </think>, <answer>, evidence XML, bbox XML, or a standalone final-answer line.
- answer is exactly one uppercase option letter such as "C".
- Do not wrap the JSON in Markdown.

Schema only; the ellipsis is not a length example:
{
  "think_content": "...",
  "answer": "C"
}
\end{promptbox}

\begin{figure*}[t!]
    \centering
    \includegraphics[width=\textwidth]{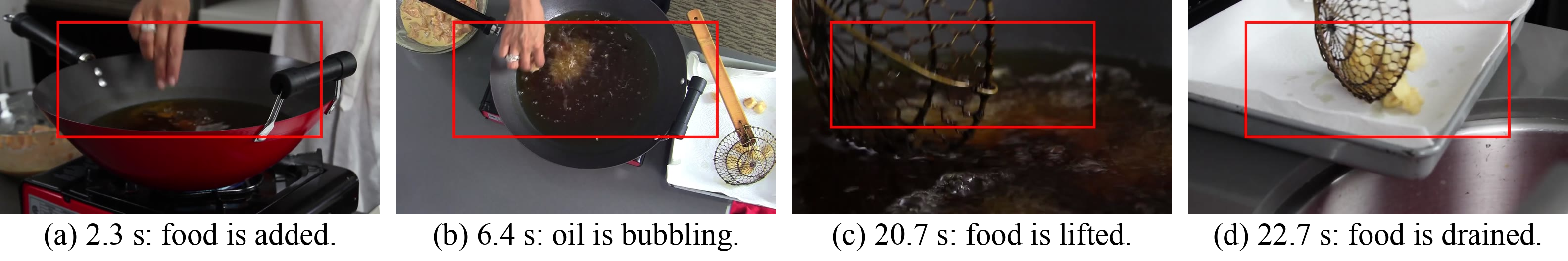}
    \caption{Localized visual regions in a representative training example.
    Red boxes mark the regions annotated in the evidence groups. Panels
    (a)--(b) form the first group, while panels (c)--(d) form the second.}
    \label{fig:training_case}
\end{figure*}

\begin{promptbox}{User Prompt}
<video>

## Task Context

Question:
{question}

Options:
{options}

## Original Reasoning

{original_reasoning_with_placeholders}

## Original Content

{original_content_with_placeholders}

## Reference Final Answer

{reference_answer}

## Evidence Catalog

{evidence_catalog}

Retain substantive observations, option analysis, evidence interpretation, hypotheses, corrections, and verification without adding filler. Before answering, privately verify that no substantive reasoning stage or self-correction was omitted.

Return only the JSON object with think_content and answer.
\end{promptbox}

In the polishing prompt above, \emph{Original Reasoning} refers to the content
inside the \texttt{think} block of the rejection-sampled response, while
\emph{Original Content} refers to the content following that block.

\section{Case Study}
\label{app:case_study}

We present a representative example from the supervised training set. The question
requires recognizing the main activity from visual evidence distributed across
the video. Figure~\ref{fig:training_case} shows the four localized visual regions used
by two evidence groups. The bounding boxes are shown in the normalized
coordinate representation used by VideoLVR.

\begin{promptbox}{Question}
<video>
What is the main activity shown in the video?
A. Boiling water
B. Deep-frying food
C. Baking a cake
D. Making a salad
\end{promptbox}

\begin{promptbox}{Evidence-Grounded Reasoning Trajectory}
<think>
The video shows a person adding pieces of food into a wok with hot oil. <evidence claim="the oil in the wok is bubbling vigorously as food is added"><box t="2.3" content="the wok with hot oil and the first piece of food being added">[0.15,0.10,0.85,0.65]</box><box t="6.4" content="the oil bubbling around the food pieces">[0.15,0.10,0.85,0.65]</box></evidence> shows that the oil bubbles vigorously around the food as it is added, which is characteristic of deep-frying.

This rules out boiling water (A) because boiling water does not produce such vigorous bubbling with solid food pieces and the oil is dark and not clear water. Baking (C) and making a salad (D) are also eliminated since there is no oven or mixing of raw vegetables. As <evidence claim="the person uses a skimmer to remove the fried food from the oil"><box t="20.7" content="a skimmer lifting the fried food from the oil">[0.10,0.10,0.80,0.60]</box><box t="22.7" content="the fried food being placed on a paper towel to drain">[0.15,0.10,0.85,0.65]</box></evidence> shows, the person uses a skimmer to remove the fried food from the oil and drain it, which is a common step in deep-frying. Therefore, the main activity is deep-frying food.

</think>
B
\end{promptbox}

This trajectory contains two evidence groups. The first uses regions at 2.3
and 6.4 seconds to ground the visual pattern of food bubbling in hot oil. The
second group uses regions at 20.7 and 22.7 seconds to ground the
subsequent removal and draining of the fried food. During training, the two
regions within each group are ordered by timestamp and concatenated into one
grouped latent target. Each evidence group is then replaced by its corresponding
latent block. Concretely, we use \texttt{<bot>} and \texttt{<eot>} as the
start and end markers, respectively, and \texttt{<latent>} for each latent
position between them. A training block of length $L_k$ therefore has the form
\texttt{<bot><latent>...<latent><eot>}. The hidden state immediately preceding
each \texttt{<latent>} position is supervised to predict the corresponding
visual token in the grouped target. During decoding, generating \texttt{<bot>}
initiates latent-state feedback; after $K$ latent inputs, the controller
inserts \texttt{<eot>} and resumes text generation. The supervised checkpoint
uses projected hidden states as feedback, whereas the reinforcement-learned
model samples visual token indices from the
visual token distribution and uses the corresponding embeddings as
feedback. These sampled indices are discrete visual actions
during RL rollout, and their likelihoods enter the TVPO
objective in Section~\ref{sec:categorical_rl}. This yields a
reasoning trajectory interleaved with two latent blocks while omitting textual
evidence annotations, including timestamps and bounding-box coordinates, from
the generation target.

\end{document}

%% file: sections/0_abstract_add_rl.tex
\begin{abstract}
Chain-of-thought reasoning has substantially advanced video understanding in multimodal large language models (MLLMs). However, language-based reasoning traces do not provide a reliable mechanism for models to revisit relevant visual evidence, leading to visual forgetting and impairing their ability to reason over video content.
Existing approaches maintain visual grounding during reasoning in two main ways: they either refer to evidence using textual coordinates, which specify symbolic locations rather than the visual content itself, or reinsert cropped regions or frames into the multimodal context, preserving visual details at the cost of substantial context expansion.
To address these limitations, we introduce \textbf{VideoLVR}, which constructs latent visual evidence from local regions at relevant video moments, thereby providing visual support for intermediate reasoning steps.
To supervise latent visual evidence generation, we introduce a visual token prediction objective that formulates visual token reconstruction as prediction over a vocabulary derived from the input video, thereby unifying visual evidence reconstruction and text generation within the same token prediction framework. This shared prediction framework enables MLLMs to reason over interleaved text tokens and latent visual states while integrating evidence from different moments.
We further extend VideoLVR to reinforcement learning by treating sampled visual tokens as actions alongside text tokens and jointly optimizing the policy using final-answer rewards.
By efficiently revisiting task-relevant visual information during reasoning, VideoLVR improves video understanding without substantially expanding the multimodal context.
Experiments on multiple video benchmarks demonstrate its effectiveness, especially on tasks requiring fine-grained visual evidence.
\end{abstract}

%% file: sections/1_intro_add_rl.tex
\section{Introduction}

Recent advances in multimodal large language models (MLLMs) have enabled increasingly complex video understanding and reasoning. Supervised chain-of-thought training and reinforcement learning have further strengthened these capabilities, allowing models to solve video questions that require temporal localization and multi-step reasoning~\citep{NEURIPS2025_8eb39768,NEURIPS2025_06b71ad9,maaz2025videor2reinforcingconsistentgrounded}. However, the resulting traces remain largely language-based. As reasoning proceeds, models may gradually lose access to task-relevant visual evidence, resulting in visual forgetting and limiting their ability to reason over video content~\citep{sun-etal-2025-mitigating-visual,sun2026thinkinghurtsmitigatingvisual}. This makes it difficult to revisit the localized and temporally sparse evidence required by later steps.

One line of work grounds reasoning with textual coordinates, using timestamps for temporal evidence and bounding boxes for spatial evidence~\citep{maaz2025videor2reinforcingconsistentgrounded,meng2026openo3videogroundedvideoreasoning,NEURIPS2025_a92ee7e4,wang2026traceableevidenceenhancedvisual}. Although compact and traceable, these coordinates encode the locations of evidence rather than its visual content. Other methods invoke visual tools to crop or zoom into image regions, or to select additional video frames and clips~\citep{zheng2026deepeyesincentivizingthinkingimages,NEURIPS2025_5c4e0e38,Zhang_2026_CVPR}. Although these tools expose fine-grained visual details, tool-augmented reasoning is limited to a predefined set of visual operations~\citep{li2025latentvisualreasoning}. Moreover, each returned crop, frame, or clip must be re-encoded and inserted into the multimodal context, causing the context to grow as visual tool calls accumulate, particularly in video reasoning~\citep{li2025latentvisualreasoning,zheng2026deepeyesincentivizingthinkingimages,Zhang_2026_CVPR}. Neither formulation provides a compact and flexible way to revisit relevant visual content while preserving its visual semantics.

Latent visual reasoning (LVR)~\citep{li2025latentvisualreasoning}, which trains an MLLM to autoregressively reconstruct question-relevant visual tokens and interleave them with textual reasoning, offers a promising alternative by representing intermediate visual evidence with continuous hidden states rather than text or additional visual observations. However, most prior latent visual reasoning methods focus on the image domain, where latent states are learned using image-based visual targets or image-level bottlenecks~\citep{Wang_2026_CVPR,Yang_2026_CVPR,Li_2026_CVPR,wu2026lavitaligninglatentvisual,miao2026gapgranularalignmentparadigm,xu2026semanticenrichedlatentvisualreasoning}. In video reasoning, a single reasoning step may require local visual evidence from several moments. Evidence from any single moment may be insufficient. The model must relate visual information across time to support the reasoning step. Meanwhile, neighboring frames often contain similar visual tokens~\citep{Fu_2025_ICCV,Tao_2025_CVPR}. MSE feature regression aligns each prediction with its target independently, without explicitly distinguishing that target from other visually similar tokens in the same video.

To this end, we introduce \textbf{VideoLVR}, which learns to generate latent visual evidence to support intermediate video reasoning. Each latent block represents evidence from local regions supporting the same reasoning step, allowing evidence from different video moments to be modeled jointly. We supervise these blocks with visual token sequences derived from spatio-temporal grounding annotations in evidence-grounded reasoning trajectories. To distinguish the target evidence from similar visual content, we introduce visual token prediction (VTP), which trains each latent query to predict its target token among all visual tokens from the input video. Together with next-token prediction for text, this objective trains VideoLVR to generate reasoning interleaved with latent visual evidence. At inference, the supervised VideoLVR model projects decoder hidden states into latent visual states and feeds them back into the model without inserting additional frames or region crops into the multimodal context. For reinforcement learning, we introduce text--visual policy optimization (TVPO), which samples visual tokens from the input video's visual vocabulary using our visual token prediction formulation and jointly optimizes text and visual token actions with final-answer rewards.
Experiments across multiple video understanding and reasoning benchmarks demonstrate its effectiveness, with particular gains on tasks requiring fine-grained visual evidence. Our main contributions are summarized as follows:
\begin{itemize}
    \item We introduce VideoLVR, which learns from evidence-grounded reasoning trajectories to construct grouped latent visual evidence from local regions across video moments, allowing temporally distributed evidence to be modeled jointly.
    \item We propose visual token prediction (VTP) over the input video's visual vocabulary, unifying latent visual evidence generation with text token prediction. Building on VTP, we introduce TVPO to jointly optimize text and visual token actions through reinforcement learning.
    \item We evaluate VideoLVR across multiple video reasoning benchmarks and conduct extensive ablations of its key designs. The results show particular gains on tasks requiring fine-grained visual evidence.
\end{itemize}

%% file: sections/2_relatedwork_add_rl.tex
\section{Related Work}

\paragraph{Video Reasoning with Explicit Evidence.}
Recent work grounds video reasoning using explicit temporal and spatial references. Video-R2 improves temporal grounding through timestamp-aware supervision and a temporal alignment reward~\citep{maaz2025videor2reinforcingconsistentgrounded}, while Open-o3-Video interleaves timestamps and bounding-box coordinates with textual reasoning~\citep{meng2026openo3videogroundedvideoreasoning}. Related image reasoning methods similarly localize relevant regions to ground intermediate reasoning~\citep{NEURIPS2025_a92ee7e4,wang2026traceableevidenceenhancedvisual,wang2026vgrvisualgroundedreasoning}. These methods make evidence traceable, but their coordinates encode symbolic locations rather than visual semantics.

\paragraph{Revisiting Visual Evidence during Reasoning.}
Long multimodal reasoning can progressively forget visual information, motivating methods that revisit relevant evidence~\citep{sun-etal-2025-mitigating-visual,sun2026thinkinghurtsmitigatingvisual}. DeepEyes and VLM-R$^3$ select and inspect image regions during reasoning~\citep{zheng2026deepeyesincentivizingthinkingimages,NEURIPS2025_5c4e0e38}. For videos, VITAL dynamically samples additional frames, while VideoSearcher combines temporal localization, spatial inspection, and multimodal search~\citep{Zhang_2026_CVPR,gao2026videosearcherempoweringvideodeep}. These methods recover fine-grained evidence by inserting regions, frames, or retrieved items as additional multimodal observations.

\paragraph{Latent Visual Reasoning.}
Continuous latent reasoning has been explored in language models, with Coconut feeding hidden states back as input embeddings to perform intermediate computation without decoding it into text~\citep{hao2025traininglargelanguagemodels}. Recent visual reasoning methods extend this idea by generating or refining latent visual representations. LVR reconstructs question-relevant visual tokens through latent states interleaved with textual reasoning~\citep{li2025latentvisualreasoning}. Other approaches learn continuous visual thoughts through visual-semantic distillation, task supervision, reinforcement learning, or latent bottlenecks~\citep{Wang_2026_CVPR,Yang_2026_CVPR,Li_2026_CVPR}. LaViT, GAP, and Semantic-Enriched Latent Visual Reasoning further improve latent supervision through attention distillation, feature-space alignment, curriculum gating, or region-centric attribute supervision~\citep{wu2026lavitaligninglatentvisual,miao2026gapgranularalignmentparadigm,xu2026semanticenrichedlatentvisualreasoning}. Diagnostic studies nevertheless show that models may bypass latent tokens, generate latents that deviate from their targets, or under-utilize their visual semantics because of shortcuts through the visual input~\citep{viveiros2026whatsholdinglatentvisual,zhang2026visuallatentsknowsay}.

Beyond continuous latent-state reasoning, Mull-Tokens uses a fixed set of modality-agnostic special tokens as a compact internal multimodal scratchpad, first aligning them with interleaved text--image traces and then relaxing training to final-answer supervision~\citep{Ray_2026_CVPR}. UniVR instead autoregressively generates visual reasoning trajectories within the discrete token space of a unified generation model, targeting physical dynamics and long-horizon planning~\citep{ren2026univrthinkingvisualspace}. VideoLatent extends latent reasoning to videos through latent injection, self-forcing, and alignment and diversity objectives trained from standard video question--answer triplets~\citep{hu2026videolatentvideolanguagelearninglatent}. However, these methods either provide generic computation slots, generate discrete visual trajectories, or learn video latents without step-level supervision from localized spatio-temporal evidence. VideoLVR addresses this gap by grouping temporally distributed local regions that support the same reasoning step and supervising latent visual evidence generation through prediction over the visual vocabulary of the input video.

%% file: sections/3_methods_add_rl.tex
\begin{figure*}[!t]
    \centering
    \includegraphics[width=\linewidth]{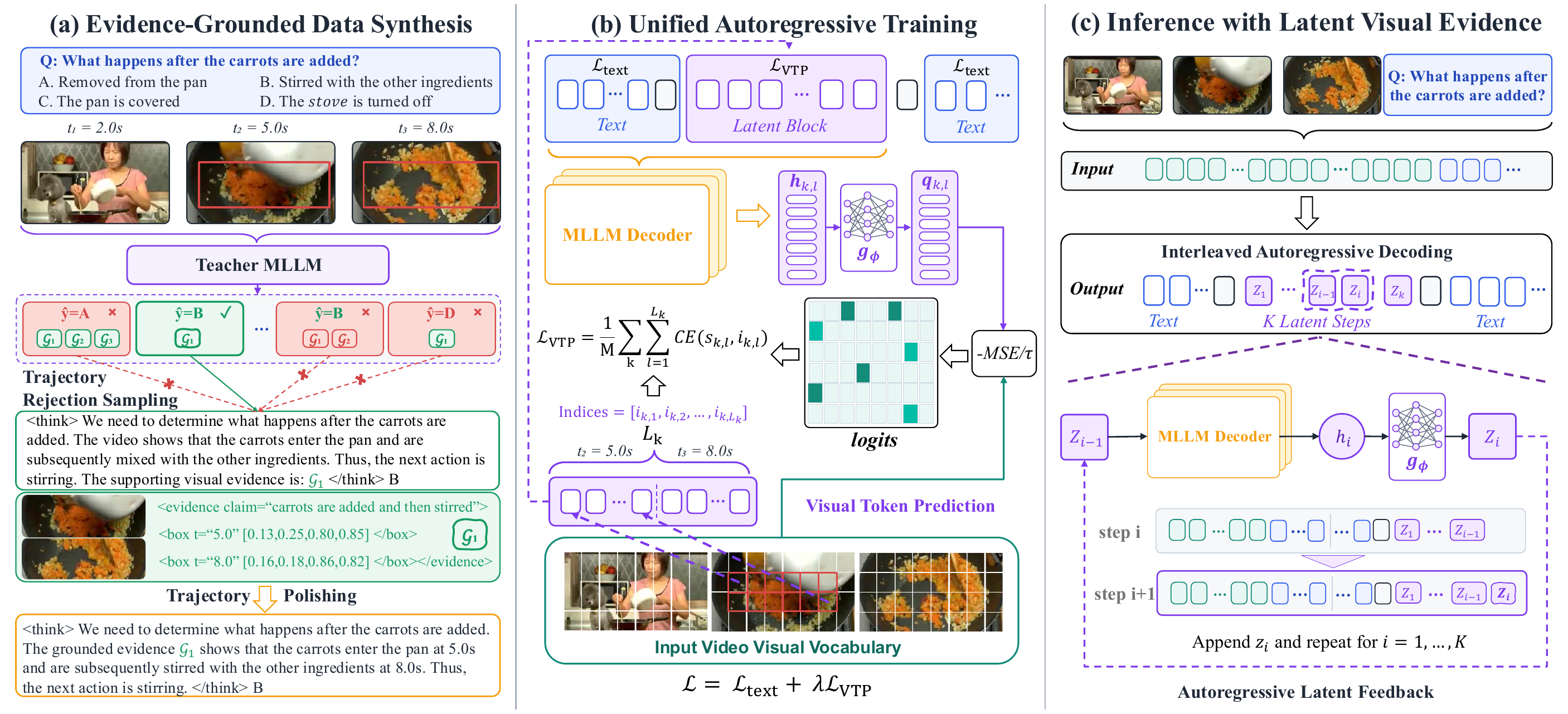}
    \caption{Overview of VideoLVR. (a) A teacher MLLM synthesizes evidence-grounded reasoning trajectories through rejection sampling and trajectory polishing. (b) Visual tokens from regions supporting the same reasoning step are grouped and ordered by timestamp to form latent targets. VideoLVR jointly trains text generation with next-token prediction and latent visual evidence generation with visual token prediction over the input-video visual vocabulary. (c) At inference with the supervised model, each decoder hidden state is projected into a latent visual state and fed back as the input for the next decoding step, enabling autoregressive reasoning over latent visual evidence without additional visual observations.}
    \label{fig:method_overview}
\end{figure*}

\section{Method}\label{sec:method}

As illustrated in Figure~\ref{fig:method_overview}, VideoLVR first synthesizes evidence-grounded reasoning trajectories and converts supporting local regions into grouped latent targets, with regions within each target ordered by timestamp. It jointly learns text generation and latent visual evidence generation through next-token prediction and visual token prediction. During inference with the supervised model, decoder hidden states are projected into latent visual states and fed back autoregressively to support subsequent reasoning without introducing additional visual observations. We next describe the construction of grounded trajectories and grouped latent targets, followed by the learning objective and autoregressive inference procedure.

\subsection{Evidence-Grounded Data Synthesis}
\label{sec:data_construction}

Starting from video question--answer data, we synthesize evidence-grounded reasoning trajectories through two stages: evidence-grounded trajectory rejection sampling and evidence-guided trajectory polishing.

\paragraph{Evidence-Grounded Trajectory Rejection Sampling.}
For each video-question pair, we prompt a teacher MLLM to generate a reasoning trajectory that ends with a final answer.
The trajectory interleaves reasoning text with spatio-temporal visual evidence groups. We denote the $k$-th evidence group by $\mathcal{G}_k$, which associates an intermediate reasoning claim $c_k$ with a sequence of localized visual regions $\mathcal{R}_k$. Each region is specified by a timestamp and a bounding box. The regions in $\mathcal{R}_k$ may come from the same video moment or span multiple moments, allowing $\mathcal{G}_k$ to capture both spatially distributed evidence within a frame and temporally distributed evidence across frames.

Formally, we represent the $k$-th evidence group $\mathcal{G}_k$ as
\begin{equation}
    \mathcal{G}_k = (c_k, \mathcal{R}_k), \qquad
    \mathcal{R}_k =
    \left(\mathbf{r}_{k,j}\right)_{j=1}^{m_k},
\end{equation}
where $c_k$ is the reasoning claim and $\mathcal{R}_k$ is the sequence of $m_k$ localized visual regions that support it. Each region is represented as $\mathbf{r}_{k,j}=(t_{k,j},b_{k,j})$, where $t_{k,j}$ is its timestamp and $b_{k,j}=(x_{k,j}^{1},y_{k,j}^{1},x_{k,j}^{2},y_{k,j}^{2})\in[0,1]^4$ specifies the normalized coordinates of the top-left and bottom-right corners. During training, each evidence group serves as a single unit for constructing a latent block.

We apply rejection sampling based on answer correctness and evidence validity. A response is retained only if its final answer is judged equivalent to the reference answer and it contains at least one well-formed evidence group, with each group specifying a reasoning claim and one or more valid localized regions. We further reject malformed timestamps and invalid or overly broad bounding boxes. Rejected responses are resampled from the teacher using the validation errors as feedback. The accepted responses form the raw evidence-grounded trajectories passed to the polishing stage.

\paragraph{Evidence-Guided Trajectory Polishing.}
Although the teacher is instructed to interleave evidence groups with the reasoning steps that use them, rejection-sampled trajectories do not always exhibit the desired ordering. Evidence groups may appear after the relevant reasoning statements or cluster near the end of the response. Directly converting these groups into latent blocks would therefore misalign the latent evidence with the reasoning steps that use it.

To correct this ordering, we provide the original video and rejection-sampled trajectory to the teacher MLLM, which rewrites the trajectory to adjust the
placement of evidence groups. The teacher preserves the original reasoning content while relocating each evidence group to the point where its visual evidence is used. We retain only trajectories that preserve the final answer, include every original evidence group exactly once, and place each group before the reasoning statement that uses the visual evidence it contains.

\subsection{Grouped Latent Visual Evidence}
\label{sec:latent_evidence}

We first map each localized visual region $\mathbf{r}_{k,j}$ to a target sequence $\mathcal{E}_{k,j}$ of visual token embeddings selected from the encoded video. Specifically, we map its timestamp $t_{k,j}$ to the corresponding temporal position in the encoded video and project its normalized bounding box $b_{k,j}$ onto the spatial token grid after visual token merging. We select the grid cells intersected by the projected box. Their visual token embeddings form $\mathcal{E}_{k,j}$, ordered from left to right within each row and from top to bottom across rows.

The evidence supporting a reasoning claim may comprise multiple localized visual regions from one or more video moments. We first order the regions in $\mathcal{G}_k$ by timestamp. We then concatenate their visual-token sequences in temporal order, without inserting any boundary or separator tokens between regions, forming a single grouped target $\widetilde{\mathcal{E}}_k$. Regions with the same timestamp retain their original annotation order. We use $\widetilde{\mathcal{E}}_k$ to supervise the latent block corresponding to claim $c_k$.

We use the special tokens \texttt{<bot>} and \texttt{<eot>} to mark the beginning and end of each latent block, respectively. During training, each evidence group is replaced by $L_k$ latent positions between these tokens, where
\begin{equation}
    L_k = \sum_{j=1}^{m_k}
    \left|\mathcal{E}_{k,j}\right|,
\end{equation}
and $\left|\mathcal{E}_{k,j}\right|$ denotes the number of visual tokens selected for region $\mathbf{r}_{k,j}$. Each hidden state immediately preceding a latent position is supervised to predict the corresponding token in $\widetilde{\mathcal{E}}_k$. Textual evidence annotations, including timestamps and bounding-box coordinates, are omitted from the generation target. This produces a training trajectory that interleaves textual reasoning with the latent visual evidence corresponding to each reasoning step. Because all supervision targets are drawn from the encoded input video, the procedure does not require cropping, re-encoding, or inserting additional visual observations into the multimodal context.

\subsection{Visual Token Prediction}\label{sec:visual_token_prediction}

Adjacent video frames often have substantial overlap in visual content, making spatially corresponding visual tokens highly similar in the feature space~\citep{Fu_2025_ICCV,Tao_2025_CVPR}. Under such temporal redundancy, MSE feature regression aligns each prediction with its target independently, without explicitly distinguishing that target from visually similar tokens in the same video. We therefore introduce a visual token prediction objective over the input-video visual vocabulary. Let $\mathcal{V}=(\mathbf{v}_1,\ldots,\mathbf{v}_N)$ denote the candidate vocabulary formed by all $N$ visual tokens encoded from the input video. The tokens localized by spatio-temporal grounding serve as prediction targets.

For the $l$-th position in the latent block constructed from $\mathcal{G}_k$, let $\mathbf{h}_{k,l}$ denote the hidden state immediately preceding that position, and let $i_{k,l}\in\{1,\ldots,N\}$ denote the vocabulary index of the corresponding target token in $\widetilde{\mathcal{E}}_k$. We implement $g_\phi$ as a two-layer MLP that maps $\mathbf{h}_{k,l}$ into the visual embedding space, producing the visual query $\mathbf{q}_{k,l}=g_\phi(\mathbf{h}_{k,l})$. The same projection head is used for visual token prediction during training and latent-state feedback during inference. We score each candidate visual token by its negative mean squared distance to the query:
\begin{equation}
    s_{k,l,i} =
    -\frac{\left\|\mathbf{q}_{k,l}-\mathbf{v}_i\right\|_2^2}{d\tau},
\end{equation}
where $d$ is the feature dimension of $\mathbf{q}_{k,l}$ and $\mathbf{v}_i$, and $\tau$ is the temperature. The resulting visual token distribution is

\begin{equation}
    p_{\mathrm{vis}}(i\mid \mathbf{q}_{k,l},\mathcal{V}) =
    \frac{\exp(s_{k,l,i})}{\sum_{n=1}^{N}\exp(s_{k,l,n})},
    \quad i=1,\ldots,N.
    \label{eq:visual_token_distribution}
\end{equation}
The visual token prediction objective is
\begin{equation}
    \mathcal{L}_{\mathrm{VTP}} =
    -\frac{1}{M}\sum_k\sum_{l=1}^{L_k}
    \log p_{\mathrm{vis}}
    \left(i_{k,l}\mid\mathbf{q}_{k,l},\mathcal{V}\right),
\end{equation}
where $M=\sum_k L_k$ is the number of latent target positions. Non-target tokens from the same video act as competing vocabulary entries, requiring each query to identify its grounded target among the visual content of that video.

During training, the target visual embeddings are used as the input embeddings at their corresponding latent positions, while the immediately preceding hidden states are optimized by $\mathcal{L}_{\mathrm{VTP}}$. This corresponds to teacher forcing over the latent sequence. The standard next-token objective $\mathcal{L}_{\mathrm{text}}$ is computed only over ordinary response tokens and \texttt{<bot>}, excluding latent positions and \texttt{<eot>}. This asymmetry follows the inference procedure: the model predicts \texttt{<bot>} to initiate latent reasoning, whereas \texttt{<eot>} is inserted by the decoding controller after $K$ latent steps, where $K$ is the fixed inference budget per block. The supervised objective is
\begin{equation}
    \mathcal{L}_{\mathrm{SFT}} = \mathcal{L}_{\mathrm{text}}
    + \lambda\mathcal{L}_{\mathrm{VTP}},
\end{equation}
where $\lambda$ balances text token prediction and visual token prediction.
Thus, textual positions are trained through discrete next-token prediction, whereas latent positions are supervised to identify grounded visual tokens within the input video's visual vocabulary,
all within the same autoregressive framework.

\subsection{Text--Visual Policy Optimization}
\label{sec:categorical_rl}

We further extend VideoLVR to reinforcement learning through text--visual policy optimization (TVPO), treating text and visual token generation as actions for joint optimization. At the $l$-th position in the $k$-th latent block, we compute the visual query $\mathbf{q}_{k,l}=g_\phi(\mathbf{h}_{k,l})$. Following Equation~\ref{eq:visual_token_distribution}, we score each token in the input-video visual vocabulary by its negative mean squared distance to the query. A temperature-scaled softmax converts these scores into visual action probabilities. We then sample a visual token action:
\begin{equation}
    i_{k,l} \sim p_{\mathrm{vis}}(\cdot\mid\mathbf{q}_{k,l},\mathcal{V}).
    \label{eq:categorical_visual_rollout}
\end{equation}
Here $i_{k,l}$ is the sampled index in the full input-video visual vocabulary $\mathcal{V}$, and $\mathbf{v}_{i_{k,l}}$ is the corresponding token embedding. In supervised training, the target index is specified by grounding annotations; during RL rollouts, the index is sampled from $p_{\mathrm{vis}}$ over the same vocabulary, and its embedding $\mathbf{v}_{i_{k,l}}$ is used as the next latent input. Each sampled vocabulary index is a discrete visual action, analogous to a generated text token. The interleaved text and visual actions form a unified action sequence that can be jointly optimized with group relative policy optimization (GRPO)~\citep{shao2024deepseekmath} using final-answer rewards.

Starting from the supervised model, we jointly optimize the language model and projection head, $\Theta=(\theta,\phi)$, using GRPO. For each video-question input $q$, we sample $G$ reasoning trajectories $\{o_i\}_{i=1}^{G}$ and assign reward $R_i=1$ if the final answer is correct and $0$ otherwise. The group-normalized advantage $A_i$ is shared by all sampled text and visual token actions in trajectory $o_i$. The TVPO objective is
\begin{equation}
\begin{aligned}
    &\mathcal{J}_{\mathrm{RL}}(\Theta)
    = \frac{1}{\mathbb{E}[\sum_{i=1}^{G}|o_i|]}
    \mathbb{E}\!\Biggl[
    \sum_{i=1}^{G}\sum_{t=1}^{|o_i|}\\
    &\quad{}\min\!\Bigl[
    r_{i,t}A_i,\,
    \operatorname{clip}(r_{i,t},1-\epsilon,1+\epsilon)A_i
    \Bigr]\Biggr].
\end{aligned}
    \label{eq:joint_text_visual_rl}
\end{equation}
Here $o_i$ is the interleaved sequence of sampled text and visual token actions, and $\epsilon$ is the clipping threshold. The policy ratio is defined as follows:

\begin{equation}
    r_{i,t}(\Theta)=
    \frac{\pi_{\Theta}(o_{i,t}\mid q,o_{i,<t})}
    {\pi_{\Theta_{\mathrm{old}}}(o_{i,t}\mid q,o_{i,<t})}.
    \label{eq:joint_policy_ratio}
\end{equation}
Here $o_{i,t}$ is the $t$-th sampled text or visual token action in $o_i$, $o_{i,<t}$ is the preceding output sequence, and $\Theta_{\mathrm{old}}$ denotes the policy parameters before the update. At textual positions, $\pi_\Theta$ uses text token probabilities. At latent positions, $o_{i,t}$ denotes the sampled vocabulary index $i_{k,l}$, and $\pi_\Theta$ uses its probability under $p_{\mathrm{vis}}$.

\subsection{Inference with Latent Visual Evidence}
\label{sec:inference}

During supervised training, the length of each latent block is determined by the
target-dependent length $L_k$ defined above, whereas each generated block uses a fixed inference-time budget of $K$ latent visual states.

VideoLVR generates text autoregressively until it emits \texttt{<bot>}. For the supervised model, the hidden state corresponding to \texttt{<bot>} is mapped by the learned projection head $g_\phi$ into the first latent visual state,
which is used as the input embedding for the next decoding step. This process is repeated autoregressively: each subsequent decoder hidden state is projected into a latent visual state and fed back as the next latent input. Once $K$ latent visual states have been generated, the decoding procedure inserts \texttt{<eot>} and resumes text generation. For the reinforcement-learned model, we instead use the embeddings of sampled visual tokens as feedback.

%% file: sections/4_exp_add_rl.tex
\section{Experiments}
\label{sec:experiments}

\subsection{Experimental Setup}
\label{sec:exp_setup}

\paragraph{Training Data.}
We construct the training data from video question--answer samples released with OneThinker~\citep{Feng_2026_CVPR} and Video-R1~\citep{NEURIPS2025_8eb39768}. Following our evidence-grounded data synthesis pipeline described in Section~\ref{sec:data_construction}, we use Qwen3-VL-30B-A3B-Thinking~\citep{bai2025qwen3vltechnicalreport} as the teacher model to generate and polish evidence-grounded reasoning trajectories in which localized spatio-temporal evidence is integrated into the reasoning process. We retain only answer-consistent trajectories after quality checks on the generated reasoning and evidence annotations. This process yields approximately 8.8K high-quality trajectories for training VideoLVR during supervised learning.
For reinforcement learning, we use 19K video multiple-choice questions from OneThinker, excluding questions that the supervised model answers correctly in all eight responses.

\paragraph{Baselines.}
We compare VideoLVR with video MLLMs that follow different reasoning paradigms. Video-R1~\citep{NEURIPS2025_8eb39768} and VideoRFT~\citep{NEURIPS2025_06b71ad9} enhance text-space video reasoning through supervised fine-tuning and reinforcement learning. OneThinker~\citep{Feng_2026_CVPR} provides a unified reasoning model for images and videos. Open-o3-Video~\citep{meng2026openo3videogroundedvideoreasoning} generates timestamps and bounding boxes within textual reasoning trajectories to provide explicit spatio-temporal grounding. VideoSearcher~\citep{gao2026videosearcherempoweringvideodeep} combines temporal localization, spatial focusing, and multimodal search through multi-tool agentic reasoning for video deep research. LVR~\citep{li2025latentvisualreasoning} reconstructs question-relevant visual features in a continuous representation space, while Mull-Tokens~\citep{Ray_2026_CVPR} uses a fixed set of modality-agnostic latent tokens as an internal multimodal scratchpad. We additionally reproduce LVR with Qwen3-VL-8B-Instruct on our 8.8K synthesized trajectories using its MSE feature regression objective and denote it as LVR-8B. We evaluate the released checkpoints of the protocol-aligned baselines using our evaluation setting. For VideoSearcher, we report results from its original evaluation setting and mark them with a dagger. For reference, we report Qwen2.5-VL-7B-Instruct~\citep{bai2025qwen25vltechnicalreport} together with the Qwen3-VL-8B-Instruct~\citep{bai2025qwen3vltechnicalreport} backbone used by VideoLVR, enabling comparison with the corresponding models before video-reasoning training.

\begin{table*}[!t]
\centering
{
\small
\setlength{\tabcolsep}{1.5pt}
\begin{tabular}{@{}lccccccccc@{}}
\toprule
\multirow{2}{*}{\textbf{Method}}
& \multicolumn{4}{c}{\textbf{Video-MMMU}}
& \multirow{2}{*}{\textbf{TempComp.}}
& \multirow{2}{*}{\textbf{MMVU}}
& \multirow{2}{*}{\textbf{LV-R.}}
& \multirow{2}{*}{\textbf{VideoMathQA}}
& \multirow{2}{*}{\textbf{Avg}} \\
\cmidrule(lr){2-5}
& \textbf{Percep.}
& \textbf{Comp.}
& \textbf{Adapt.}
& \textbf{Avg}
& & & & & \\
\midrule

Qwen2.5-VL-7B-Instruct
& 60.00
& \underline{44.00}
& 24.00
& 42.67
& 61.97
& \underline{53.40}
& \underline{71.20}
& 22.62
& 50.37 \\

Video-R1-7B
& 45.33
& 37.67
& 29.00
& 37.33
& 60.83
& 45.90
& 59.30
& 20.00
& 44.67 \\

VideoRFT-7B
& 49.33
& 39.33
& 25.67
& 38.11
& 59.76
& 49.10
& 64.00
& 23.10
& 46.81 \\

Open-o3-Video-7B
& \underline{60.33}
& 41.33
& \textbf{35.33}
& \underline{45.67}
& \textbf{72.20}
& 51.00
& 63.60
& \underline{24.41}
& \underline{51.38} \\

Mull-Tokens-7B
& 51.33
& 39.67
& 29.00
& 40.00
& 57.41
& 29.20
& 61.40
& 16.90
& 40.98 \\

LVR-7B
& 53.93
& 40.44
& 24.00
& 39.46
& 56.98
& 46.60
& 56.80
& 23.10
& 44.59 \\

\textbf{VideoLVR-7B}
& \textbf{61.33}
& \textbf{46.67}
& \underline{30.67}
& \textbf{46.22}
& \underline{69.37}
& \textbf{54.50}
& \textbf{73.80}
& \textbf{24.52}
& \textbf{53.68} \\

\midrule

Qwen3-VL-8B-Instruct
& \underline{79.33}
& 67.33
& 46.33
& 64.33
& 73.89
& 59.30
& 76.40
& 32.62
& 61.31 \\

OneThinker-8B
& 75.00
& 60.67
& 39.67
& 58.44
& 70.44
& 57.50
& 74.10
& 30.00
& 58.10 \\

VideoSearcher-8B$^\dagger$
& \textemdash{}
& \textemdash{}
& \textemdash{}
& 66.33
& 73.14
& 70.40
& \textemdash{}
& 31.67
& \textemdash{} \\

LVR-8B
& 78.67
& \underline{72.67}
& \underline{50.50}
& 67.28
& 76.42
& 59.50
& 76.80
& 33.57
& 62.71 \\

VideoLVR-8B
& 79.00
& \textbf{74.00}
& 49.33
& \underline{67.44}
& \underline{77.39}
& \textbf{61.70}
& \underline{78.20}
& \underline{35.00}
& \underline{63.95} \\

\textbf{VideoLVR-8B-RL}
& \textbf{81.00}
& 72.33
& \textbf{51.33}
& \textbf{68.22}
& \textbf{77.60}
& \underline{60.50}
& \textbf{79.90}
& \textbf{35.71}
& \textbf{64.39} \\

\bottomrule
\end{tabular}
}
\caption{
Main results on five video reasoning benchmarks.
VideoLVR-8B-RL denotes VideoLVR-8B after training with TVPO.
Percep., Comp., and Adapt. denote the Perception, Comprehension, and Adaptation
dimensions of Video-MMMU. TempComp. and LV-R. abbreviate TempCompass
and LongVideo-Reason, respectively. Bold indicates the best reproduced result
within each backbone family, while underlining denotes the second best.
$^\dagger$ denotes results reported under the original evaluation settings,
which are excluded from ranking.
}
\label{tab:main_results}
\end{table*}

\paragraph{Implementation Details.}
We first jointly fine-tune the language model and projection head $g_{\phi}$ while freezing the vision encoder and multimodal projector. For both training and evaluation, we sample videos at 2 FPS, retain at most 256 frames, and set the per-frame pixel budget to 131,072. Training uses a global batch size of 64 and a learning rate of $1\times10^{-5}$ with cosine decay and a 0.1 warmup ratio. We train for two epochs. Unless otherwise stated, we use $\lambda=0.1$, $\tau=0.07$, and an inference-time latent-step budget of $K=8$.
RL uses 64 prompts per batch, eight rollouts per prompt, a learning rate of $1\times10^{-6}$, and a maximum response length of 4,096. Final-answer correctness is the only reward.

\paragraph{Benchmarks and Evaluation.}
We evaluate VideoLVR on five complementary video reasoning benchmarks. Video-MMMU~\citep{hu-etal-2026-video} assesses how models acquire and use knowledge from multidisciplinary professional videos across Perception, Comprehension, and Adaptation. TempCompass~\citep{liu-etal-2024-tempcompass} evaluates temporal perception across multiple temporal aspects and task formats while reducing reliance on static visual cues. MMVU~\citep{Zhao_2025_CVPR} measures expert-level, knowledge-intensive reasoning over specialized videos from multiple disciplines. LongVideo-Reason~\citep{chen2025scalingrllongvideos} focuses on reasoning over long videos from diverse real-world domains. VideoMathQA~\citep{rasheed2025videomathqabenchmarkingmathematicalreasoning} evaluates temporally extended mathematical reasoning that integrates visual, textual, and audio information from instructional videos.

We conduct all reproduced evaluations within the LMMs-Eval framework~\citep{zhang-etal-2025-lmms} and report accuracy following the standard protocol of each benchmark; Video-MMMU further reports Perception, Comprehension, and Adaptation. Evaluation is zero-shot and allows at most 16,384 newly generated tokens. Unless otherwise stated, all reproduced results are averaged over three complete evaluations. The overall average is the unweighted mean of Video-MMMU Avg, TempCompass, MMVU, LongVideo-Reason, and VideoMathQA.

\subsection{Main Results}
\label{sec:main_results}

As shown in Table~\ref{tab:main_results}, VideoLVR-8B achieves an average score of 63.95, outperforming Qwen3-VL-8B-Instruct by 2.64 points. It improves over the backbone on every benchmark, indicating that the overall gain is not driven by a single task. As a more direct latent reasoning baseline, LVR-8B applies the MSE feature regression objective of LVR to Qwen3-VL-8B-Instruct and is trained on our 8.8K synthesized trajectories. VideoLVR-8B surpasses this reproduction by 1.24 points on average and outperforms the reproduced OneThinker-8B on all five benchmarks, demonstrating consistent advantages across different video reasoning tasks.

The most pronounced gains are on TempCompass and Video-MMMU, with improvements of 3.50 and 3.11 points, respectively. On Video-MMMU, the largest improvement comes from Comprehension, where VideoLVR gains 6.67 points. This pattern aligns with the design of VideoLVR: during training, spatio-temporal annotations ground latent supervision in video moments and local regions, while continuous latent visual states are trained to represent the visual semantics for intermediate reasoning. VideoLVR improves MMVU, LongVideo-Reason, and VideoMathQA, suggesting that its benefits extend beyond temporal perception to expert video understanding, long-range evidence integration, and video-grounded mathematical reasoning.

On Qwen2.5-VL-7B-Instruct, VideoLVR-7B raises the average score from 50.37 to 53.68 and achieves the highest average among the reproduced protocol-aligned methods based on Qwen2.5-VL-7B-Instruct. It outperforms Open-o3-Video and LVR by 2.30 and 9.09 points on average, respectively. Compared with LVR-7B, the gains are particularly pronounced on LongVideo-Reason and TempCompass. These results show that the improvements brought by VideoLVR are not limited to a particular base model.
After RL, VideoLVR-8B-RL achieves an average of 64.39. Compared with VideoLVR-8B, its average improves by 0.44 points.

\subsection{Ablation Studies}\label{sec:ablations}

\begin{table}[t]
\centering
{
\small
\setlength{\tabcolsep}{0.7mm}
\begin{tabular}{@{}lcccccc@{}}
\toprule
\textbf{Loss} & \textbf{V-MMMU} & \textbf{TC} & \textbf{MMVU} & \textbf{LV-R.} & \textbf{VMQA} & \textbf{Avg} \\
\midrule
MSE & 67.28 & 76.42 & 59.50 & 76.80 & 33.57 & 62.71 \\
VTP & \textbf{67.44} & \textbf{77.39} & \textbf{61.70} & \textbf{78.20} & \textbf{35.00} & \textbf{63.95} \\
\bottomrule
\end{tabular}
}
\caption{Comparison of MSE and visual token prediction (VTP) for latent supervision. V-MMMU, TC, LV-R., and VMQA abbreviate Video-MMMU, TempCompass, LongVideo-Reason, and VideoMathQA, respectively.}
\label{tab:vtp_objective_ablation}
\end{table}

\begin{figure*}[!t]
\centering
\includegraphics[width=0.98\textwidth]{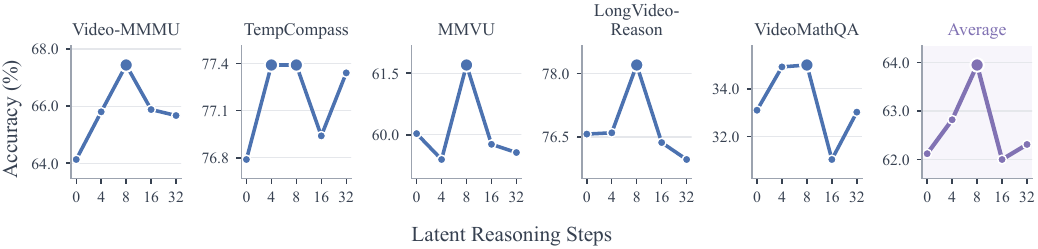}
\caption{Effect of the inference-time latent-step budget on five video benchmarks. Enlarged markers denote the best results.}
\label{fig:latent_step_ablation}
\end{figure*}

\paragraph{Effect of Visual Token Prediction.}
Table~\ref{tab:vtp_objective_ablation} compares the two latent supervision objectives under the same backbone, training data, and evaluation protocol. Replacing MSE feature regression with visual token prediction improves the average across the five benchmarks from 62.71 to 63.95, yielding a gain of 1.24 points. The improvements are largest on MMVU, VideoMathQA, and LongVideo-Reason, with gains of 2.20, 1.43, and 1.40 points, respectively. This advantage is particularly relevant under temporal redundancy because MSE aligns each latent prediction with its target independently, whereas VTP requires it to distinguish the grounded target from other visually similar tokens in the same video. The overall gain suggests that visual token prediction is more effective than MSE feature regression in improving video understanding performance.

\begin{table}[t]
\centering
{
\small
\setlength{\tabcolsep}{0.7mm}
\begin{tabular}{@{}lcccccc@{}}
\toprule
\textbf{Data} & \textbf{V-MMMU} & \textbf{TC} & \textbf{MMVU} & \textbf{LV-R.} & \textbf{VMQA} & \textbf{Avg} \\
\midrule
Standard RS & 64.89 & 75.15 & \textbf{63.60} & 77.60 & 31.43 & 62.53 \\
Ours & \textbf{67.44} & \textbf{77.39} & 61.70 & \textbf{78.20} & \textbf{35.00} & \textbf{63.95} \\
\bottomrule
\end{tabular}
}
\caption{Comparison of trajectories from standard rejection sampling and our evidence-grounded synthesis pipeline.}
\label{tab:evidence_data_ablation}
\end{table}

\paragraph{Effect of Evidence-Grounded Training Data.}
We apply standard rejection sampling to the training samples and train Qwen3-VL-8B-Instruct on the resulting reasoning trajectories. Unlike our evidence-grounded training data, these trajectories do not contain localized spatio-temporal evidence. As shown in Table~\ref{tab:evidence_data_ablation}, training on evidence-grounded reasoning trajectories improves the average from 62.53 to 63.95. The largest gains appear on VideoMathQA, Video-MMMU, and TempCompass, where the scores increase by 3.57, 2.55, and 2.24 points, respectively. The improvement on LongVideo-Reason is smaller, while MMVU decreases by 1.90 points, showing that the benefit is not uniform across tasks. Overall, the results suggest that associating intermediate reasoning with localized spatio-temporal evidence generally improves video understanding over training on reasoning trajectories obtained through standard rejection sampling.

\paragraph{Effect of Grouped Latent Visual Evidence.}
We compare three ways of organizing the localized regions associated with each evidence group. First-Region retains only the first annotated region in each group. Per-Region preserves all regions but assigns each region to a separate latent block. Grouped combines all regions into a single temporally ordered latent block, which is the strategy adopted in VideoLVR. As shown in Table~\ref{tab:evidence_organization_ablation}, Grouped achieves the highest average score of 63.95, exceeding First-Region and Per-Region by 0.80 and 0.84 points, respectively. These results suggest that retaining only one region may discard complementary temporal evidence, whereas assigning related regions to separate latent blocks can fragment evidence that jointly supports the same reasoning step. Grouping temporally distributed regions within one latent block better preserves their shared reasoning role. This result further validates the effectiveness of our synthesized training data grounded in localized spatio-temporal evidence.

\begin{table}[t]
\centering
{
\small
\setlength{\tabcolsep}{1.4pt}
\begin{tabular}{@{}lcccccc@{}}
\toprule
\textbf{Setting} & \textbf{V-MMMU} & \textbf{TC} & \textbf{MMVU} & \textbf{LV-R.} & \textbf{VMQA} & \textbf{Avg} \\
\midrule
First-Region & 66.89 & 76.76 & 60.70 & 76.90 & 34.52 & 63.15 \\
Per-Region & 66.56 & 77.05 & 60.10 & 77.30 & 34.52 & 63.11 \\
Grouped & \textbf{67.44} & \textbf{77.39} & \textbf{61.70} & \textbf{78.20} & \textbf{35.00} & \textbf{63.95} \\
\bottomrule
\end{tabular}
}
\caption{Comparison of strategies for organizing localized visual regions into latent blocks.}
\label{tab:evidence_organization_ablation}
\end{table}

\paragraph{Effect of TVPO.}
Table~\ref{tab:rl_rollout_ablation} compares the two RL variants using the same supervised VideoLVR-8B initialization and RL data. Our GRPO baseline uses continuous latent feedback and optimizes text actions only, whereas TVPO jointly optimizes text and visual actions. TVPO improves the five-benchmark average from 63.31 to 64.39, a gain of 1.08 points, and outperforms this baseline on all five benchmarks.

\begin{table}[t]
\centering
{
\small
\setlength{\tabcolsep}{0.7mm}
\begin{tabular}{@{}lcccccc@{}}
\toprule
\textbf{Method} & \textbf{V-MMMU} & \textbf{TC} & \textbf{MMVU} & \textbf{LV-R.} & \textbf{VMQA} & \textbf{Avg} \\
\midrule
GRPO & 67.11 & 77.33 & 59.40 & 77.70 & 35.00 & 63.31 \\
TVPO & \textbf{68.22} & \textbf{77.60} & \textbf{60.50} & \textbf{79.90} & \textbf{35.71} & \textbf{64.39} \\
\bottomrule
\end{tabular}
}
\caption{Comparison of GRPO and TVPO, both initialized from the same supervised VideoLVR-8B model.}
\label{tab:rl_rollout_ablation}
\end{table}

\subsection{Additional Analysis}\label{sec:latent_analysis}

\begin{figure}[t]
    \centering
    \includegraphics[width=\columnwidth]{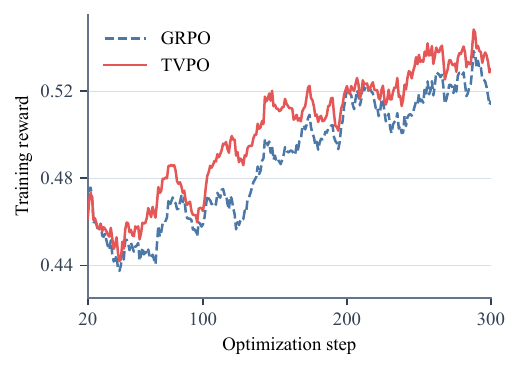}
    \caption{Comparison of training reward curves for GRPO and TVPO,
    shown as trailing 20-step moving averages.}
    \label{fig:rl_reward_comparison}
\end{figure}

\paragraph{Effect of the Inference-Time Latent-Step Budget.}
Figure~\ref{fig:latent_step_ablation} varies the inference-time latent-step budget from 0 to 32 while keeping the trained model fixed. Disabling latent visual state generation ($K=0$) yields an average of 62.12, 1.83 points below the 63.95 achieved by $K=8$, with lower performance on all five benchmarks. Allocating four latent steps recovers part of this gap, raising the average to 62.82, while eight steps provide the strongest overall result. Increasing the budget further to 16 or 32 reduces the average to 62.00 and 62.31, respectively. At the benchmark level, $K=8$ performs best on Video-MMMU, MMVU, LongVideo-Reason, and VideoMathQA, and ties $K=4$ on TempCompass. These results demonstrate that latent visual states contribute to performance, while also showing that their benefit depends on a moderate inference-time latent-step budget rather than simply increasing the number of latent steps.

\paragraph{RL Training Dynamics.}
Figure~\ref{fig:rl_reward_comparison} compares the training reward curves of GRPO and TVPO on our 19K RL dataset. Both methods show an overall increase in reward after the initial fluctuations. TVPO achieves higher rewards over most training steps and maintains this advantage in the later stages. Since the reward measures final-answer correctness, higher rewards correspond to a larger proportion of correct responses among the sampled training trajectories. These results provide training-time evidence for the effectiveness of TVPO relative to the GRPO baseline.

\paragraph{Effect of Vocabulary Temperature.}
Table~\ref{tab:vtp_temperature_ablation} examines the effect of the temperature used for visual token prediction. Performance improves steadily as $\tau$ increases from 0.01 to 0.07, with the average rising from 59.98 to 63.95, but declines at 0.10 and 0.15. The $\tau=0.07$ setting also achieves the best result on all five benchmarks. Since $\tau$ controls the sharpness of visual token discrimination, these results suggest that an overly small value produces excessively hard supervision, whereas a larger value weakens the discrimination objective; $\tau=0.07$ provides a more effective balance.

\begin{table}[t]
\centering
{
\small
\setlength{\tabcolsep}{1mm}
\begin{tabular}{@{}lcccccc@{}}
\toprule
\textbf{$\tau$} & \textbf{V-MMMU} & \textbf{TC} & \textbf{MMVU} & \textbf{LV-R.} & \textbf{VMQA} & \textbf{Avg} \\
\midrule
0.01 & 62.66 & 73.75 & 57.70 & 75.97 & 29.84 & 59.98 \\
0.03 & 65.04 & 75.64 & 59.27 & 77.95 & 30.24 & 61.63 \\
0.07 & \textbf{67.44} & \textbf{77.39} & \textbf{61.70} & \textbf{78.20} & \textbf{35.00} & \textbf{63.95} \\
0.10 & 66.00 & 76.59 & 59.53 & 77.10 & 34.52 & 62.75 \\
0.15 & 66.35 & 77.09 & 60.30 & 76.37 & 32.78 & 62.58 \\
\bottomrule
\end{tabular}
}
\caption{Ablation of the visual token prediction temperature $\tau$ on the five main benchmarks.}
\label{tab:vtp_temperature_ablation}
\end{table}

\paragraph{Effect of Latent-State Content.}
To further investigate the role of latent-state content in VideoLVR, we intervene on the feedback vector at every latent step while keeping the
eight-step latent budget and boundary markers unchanged. Zero replaces each feedback vector with a zero vector. Noise replaces it with a random vector whose $\ell_2$ norm matches that of the original feedback vector, preserving its magnitude while removing its learned content. As shown in Table~\ref{tab:latent_content_intervention}, both interventions degrade performance across all five benchmarks, reducing the average score from 63.95 to 62.75 and 62.07, respectively. These results indicate that the information encoded in the generated latent states contributes to VideoLVR's performance.

\begin{table}[t]
\centering
{
\small
\setlength{\tabcolsep}{1pt}
\begin{tabular}{@{}lcccccc@{}}
\toprule
\textbf{Setting} & \textbf{V-MMMU} & \textbf{TC} & \textbf{MMVU} & \textbf{LV-R.} & \textbf{VMQA} & \textbf{Avg} \\
\midrule
Zero & 65.11 & 76.81 & 61.00 & 76.30 & 34.52 & 62.75 \\
Noise & 65.00 & 76.86 & 60.70 & 76.10 & 31.67 & 62.07 \\
Ours & \textbf{67.44} & \textbf{77.39} & \textbf{61.70} & \textbf{78.20} & \textbf{35.00} & \textbf{63.95} \\
\bottomrule
\end{tabular}
}
\caption{Effect of interventions on latent states during inference.}
\label{tab:latent_content_intervention}
\end{table}

%% file: sections/5_conclusion_add_rl.tex
\section{Conclusion}

In this paper, we introduce VideoLVR, a framework that grounds video reasoning with latent visual evidence constructed from temporally distributed local regions. VideoLVR groups regions according to the reasoning step they support and trains the model to generate the corresponding latent visual states through visual token prediction, thereby interleaving textual reasoning with latent visual evidence in a shared autoregressive process. We further extend VideoLVR to reinforcement learning through TVPO, jointly optimizing text and visual token actions with GRPO. Across five video reasoning benchmarks and two backbone models, VideoLVR consistently improves over its base models. Controlled analyses support the benefits of evidence-grounded training data, grouped evidence organization, and visual token prediction, while latent-state interventions further show that the model benefits from the information carried by the generated latent states. Together, these results indicate that localized video evidence can be represented and used during reasoning without inserting additional visual observations into the multimodal context. A remaining limitation is our reliance on synthesized spatio-temporal grounding annotations. Future work could broaden the representational scope of latent visual states beyond localized evidence, enabling them to encode higher-level semantic information.

%% file: aaai2027.bib
@inproceedings{NEURIPS2025_06b71ad9,
 author = {Wang, Qi and Yu, Yanrui and Yuan, Ye and Mao, Rui and Zhou, Tianfei},
 booktitle = {Advances in Neural Information Processing Systems},
 editor = {D. Belgrave and C. Zhang and H. Lin and R. Pascanu and P. Koniusz and M. Ghassemi and N. Chen},
 pages = {4350--4376},
 publisher = {Curran Associates, Inc.},
 title = {VideoRFT: Incentivizing Video Reasoning Capability in MLLMs via Reinforced Fine-Tuning},
 url = {https://proceedings.neurips.cc/paper_files/paper/2025/file/06b71ad997f7e3e4b2e2f2ea12e5a759-Paper-Conference.pdf},
 volume = {38},
 year = {2025}
}

@inproceedings{NEURIPS2025_a92ee7e4,
 author = {Fan, Yue and He, Xuehai and Yang, Diji and Zheng, Kaizhi and Kuo, Ching-Chen and Zheng, Yuting and Guan, Xinze and Wang, Xin},
 booktitle = {Advances in Neural Information Processing Systems},
 editor = {D. Belgrave and C. Zhang and H. Lin and R. Pascanu and P. Koniusz and M. Ghassemi and N. Chen},
 pages = {116522--116543},
 publisher = {Curran Associates, Inc.},
 title = {GRIT: Teaching MLLMs to Think with Images},
 url = {https://proceedings.neurips.cc/paper_files/paper/2025/file/a92ee7e49a25412add0883454da6bed9-Paper-Conference.pdf},
 volume = {38},
 year = {2025}
}

@inproceedings{NEURIPS2025_5c4e0e38,
 author = {Jiang, Chaoya and Heng, Yongrui and Ye, Wei and Xu, Haiyang and Yan, Ming and Zhang, Ji and Huang, Fei and Zhang, Shikun},
 booktitle = {Advances in Neural Information Processing Systems},
 editor = {D. Belgrave and C. Zhang and H. Lin and R. Pascanu and P. Koniusz and M. Ghassemi and N. Chen},
 pages = {63841--63869},
 publisher = {Curran Associates, Inc.},
 title = {VLM-R{$^3$}: Region Recognition, Reasoning, and Refinement for Enhanced Multimodal Chain-of-Thought},
 url = {https://proceedings.neurips.cc/paper_files/paper/2025/file/5c4e0e38691e2aa08bba4cefc4c6e852-Paper-Conference.pdf},
 volume = {38},
 year = {2025}
}

@InProceedings{Zhang_2026_CVPR,
    author    = {Zhang, Haoji and Gu, Xin and Li, Jiawen and Ma, Chixiang and Bai, Sule and Zhang, Chubin and Zhang, Bowen and Zhou, Zhichao and He, Dongliang and Tang, Yansong},
    title     = {Thinking With Videos: Multimodal Tool-Augmented Reinforcement Learning for Long Video Reasoning},
    booktitle = {Proceedings of the IEEE/CVF Conference on Computer Vision and Pattern Recognition (CVPR)},
    month     = {June},
    year      = {2026},
    pages     = {32903-32914}
}

@InProceedings{Li_2026_CVPR,
    author    = {Li, Kelvin and Shang, Chuyi and Karlinsky, Leonid and Feris, Rogerio and Darrell, Trevor and Herzig, Roei},
    title     = {Latent Implicit Visual Reasoning},
    booktitle = {Proceedings of the IEEE/CVF Conference on Computer Vision and Pattern Recognition (CVPR)},
    month     = {June},
    year      = {2026},
    pages     = {33457-33466}
}

@misc{meng2026openo3videogroundedvideoreasoning,
      title={Open-o3-Video: Grounded Video Reasoning with Explicit Spatio-Temporal Evidence}, 
      author={Jiahao Meng and Xiangtai Li and Haochen Wang and Yue Tan and Tao Zhang and Lingdong Kong and Yunhai Tong and Anran Wang and Zhiyang Teng and Yujing Wang and Zhuochen Wang},
      year={2026},
      eprint={2510.20579},
      archivePrefix={arXiv},
      primaryClass={cs.CV},
      url={https://arxiv.org/abs/2510.20579}, 
}

@misc{wang2026traceableevidenceenhancedvisual,
      title={Traceable Evidence Enhanced Visual Grounded Reasoning: Evaluation and Methodology}, 
      author={Haochen Wang and Xiangtai Li and Zilong Huang and Anran Wang and Jiacong Wang and Tao Zhang and Jiani Zheng and Sule Bai and Zijian Kang and Jiashi Feng and Zhuochen Wang and Zhaoxiang Zhang},
      year={2026},
      eprint={2507.07999},
      archivePrefix={arXiv},
      primaryClass={cs.CV},
      url={https://arxiv.org/abs/2507.07999}, 
}

@misc{zheng2026deepeyesincentivizingthinkingimages,
      title={DeepEyes: Incentivizing "Thinking with Images" via Reinforcement Learning}, 
      author={Ziwei Zheng and Michael Yang and Jack Hong and Chenxiao Zhao and Guohai Xu and Le Yang and Chao Shen and Xing Yu},
      year={2026},
      eprint={2505.14362},
      archivePrefix={arXiv},
      primaryClass={cs.CV},
      url={https://arxiv.org/abs/2505.14362}, 
}

@misc{li2025latentvisualreasoning,
      title={Latent Visual Reasoning}, 
      author={Bangzheng Li and Ximeng Sun and Jiang Liu and Ze Wang and Jialian Wu and Xiaodong Yu and Hao Chen and Emad Barsoum and Muhao Chen and Zicheng Liu},
      year={2025},
      eprint={2509.24251},
      archivePrefix={arXiv},
      primaryClass={cs.CV},
      url={https://arxiv.org/abs/2509.24251}, 
}

@misc{wang2026vgrvisualgroundedreasoning,
      title={VGR: Visual Grounded Reasoning}, 
      author={Jiacong Wang and Zijian Kang and Haochen Wang and Haiyong Jiang and Jiawen Li and Bohong Wu and Ya Wang and Jiao Ran and Xiao Liang and Chao Feng and Jun Xiao},
      year={2026},
      eprint={2506.11991},
      archivePrefix={arXiv},
      primaryClass={cs.CV},
      url={https://arxiv.org/abs/2506.11991}, 
}

@misc{xu2026semanticenrichedlatentvisualreasoning,
      title={Semantic-Enriched Latent Visual Reasoning}, 
      author={Tianrun Xu and Yue Sun and Qixun Wang and Jingyi Lu and Yuan Wang and Tianren Zhang and Longteng Guo and Fengyun Rao and Jing Lyu and Feng Chen and Jing Liu},
      year={2026},
      eprint={2605.19342},
      archivePrefix={arXiv},
      primaryClass={cs.CV},
      url={https://arxiv.org/abs/2605.19342}, 
}

@misc{maaz2025videor2reinforcingconsistentgrounded,
      title={Video-R2: Reinforcing Consistent and Grounded Reasoning in Multimodal Language Models}, 
      author={Muhammad Maaz and Hanoona Rasheed and Fahad Shahbaz Khan and Salman Khan},
      year={2025},
      eprint={2511.23478},
      archivePrefix={arXiv},
      primaryClass={cs.CV},
      url={https://arxiv.org/abs/2511.23478}, 
}

@misc{miao2026gapgranularalignmentparadigm,
      title={Fill the GAP: A Granular Alignment Paradigm for Visual Reasoning in Multimodal Large Language Models}, 
      author={Yanting Miao and Yutao Sun and Dexin Wang and Mengyu Zhou and Pascal Poupart and Lei Lv and Qi Zhao and Li Wang and Hao Li and Xiaoxi Jiang and Guanjun Jiang},
      year={2026},
      eprint={2605.12374},
      archivePrefix={arXiv},
      primaryClass={cs.CV},
      url={https://arxiv.org/abs/2605.12374}, 
}

@misc{hu2026videolatentvideolanguagelearninglatent,
      title={VideoLatent: Video-Language Learning via Latent Self-Forcing}, 
      author={Zi-Yuan Hu and Zicong Tang and Shijia Huang and Yanyang Li and Michael R. Lyu and Liwei Wang},
      year={2026},
      eprint={2606.22870},
      archivePrefix={arXiv},
      primaryClass={cs.CV},
      url={https://arxiv.org/abs/2606.22870}, 
}

@misc{viveiros2026whatsholdinglatentvisual,
      title={What's Holding Back Latent Visual Reasoning?}, 
      author={André G. Viveiros and Nuno Gonçalves and André F. T. Martins and Matthias Lindemann},
      year={2026},
      eprint={2605.18445},
      archivePrefix={arXiv},
      primaryClass={cs.CV},
      url={https://arxiv.org/abs/2605.18445}, 
}

@misc{zhang2026visuallatentsknowsay,
      title={Visual Latents Know More Than They Say: Unsilencing Latent Reasoning in MLLMs}, 
      author={Xin Zhang and Qiqi Tao and Jiawei Du and Moyun Liu and Joey Tianyi Zhou},
      year={2026},
      eprint={2605.02735},
      archivePrefix={arXiv},
      primaryClass={cs.LG},
      url={https://arxiv.org/abs/2605.02735}, 
}

@InProceedings{Wang_2026_CVPR,
    author    = {Wang, Qixun and Shi, Yang and Wang, Yifei and Zhang, Yuanxing and Wan, Pengfei and Gai, Kun and Ying, Xianghua and Wang, Yisen},
    title     = {Monet: Reasoning in Latent Visual Space Beyond Image and Language},
    booktitle = {Proceedings of the IEEE/CVF Conference on Computer Vision and Pattern Recognition (CVPR)},
    month     = {June},
    year      = {2026},
    pages     = {12030-12040}
}

@InProceedings{Yang_2026_CVPR,
    author    = {Yang, Zeyuan and Yu, Xueyang and Chen, Delin and Shen, Maohao and Gan, Chuang},
    title     = {Machine Mental Imagery: Empower Multimodal Reasoning with Latent Visual Tokens},
    booktitle = {Proceedings of the IEEE/CVF Conference on Computer Vision and Pattern Recognition (CVPR)},
    month     = {June},
    year      = {2026},
    pages     = {33510-33520}
}

@misc{hao2025traininglargelanguagemodels,
      title={Training Large Language Models to Reason in a Continuous Latent Space}, 
      author={Shibo Hao and Sainbayar Sukhbaatar and DiJia Su and Xian Li and Zhiting Hu and Jason Weston and Yuandong Tian},
      year={2025},
      eprint={2412.06769},
      archivePrefix={arXiv},
      primaryClass={cs.CL},
      url={https://arxiv.org/abs/2412.06769}, 
}

@misc{wu2026lavitaligninglatentvisual,
      title={LaViT: Aligning Latent Visual Thoughts for Multi-modal Reasoning}, 
      author={Linquan Wu and Tianxiang Jiang and Yifei Dong and Haoyu Yang and Fengji Zhang and Shichaang Meng and Ai Xuan and Linqi Song and Jacky Keung},
      year={2026},
      eprint={2601.10129},
      archivePrefix={arXiv},
      primaryClass={cs.CV},
      url={https://arxiv.org/abs/2601.10129}, 
}

@inproceedings{NEURIPS2025_8eb39768,
 author = {Feng, Kaituo and Gong, Kaixiong and Li, Bohao and Guo, Zonghao and Wang, Yibing and Peng, Tianshuo and Wu, Junfei and Zhang, Xiaoying and Wang, Benyou and Yue, Xiangyu},
 booktitle = {Advances in Neural Information Processing Systems},
 editor = {D. Belgrave and C. Zhang and H. Lin and R. Pascanu and P. Koniusz and M. Ghassemi and N. Chen},
 pages = {99114--99137},
 publisher = {Curran Associates, Inc.},
 title = {Video-R1: Reinforcing Video Reasoning in MLLMs},
 url = {https://proceedings.neurips.cc/paper_files/paper/2025/file/8eb3976840e08b80dda9667562574246-Paper-Conference.pdf},
 volume = {38},
 year = {2025}
}

@misc{bai2025qwen25vltechnicalreport,
      title={Qwen2.5-VL Technical Report}, 
      author={Shuai Bai and Keqin Chen and Xuejing Liu and Jialin Wang and Wenbin Ge and Sibo Song and Kai Dang and Peng Wang and Shijie Wang and Jun Tang and Humen Zhong and Yuanzhi Zhu and Mingkun Yang and Zhaohai Li and Jianqiang Wan and Pengfei Wang and Wei Ding and Zheren Fu and Yiheng Xu and Jiabo Ye and Xi Zhang and Tianbao Xie and Zesen Cheng and Hang Zhang and Zhibo Yang and Haiyang Xu and Junyang Lin},
      year={2025},
      eprint={2502.13923},
      archivePrefix={arXiv},
      primaryClass={cs.CV},
      url={https://arxiv.org/abs/2502.13923}, 
}

@misc{bai2025qwen3vltechnicalreport,
      title={Qwen3-VL Technical Report}, 
      author={Shuai Bai and Yuxuan Cai and Ruizhe Chen and Keqin Chen and Xionghui Chen and Zesen Cheng and Lianghao Deng and Wei Ding and Chang Gao and Chunjiang Ge and Wenbin Ge and Zhifang Guo and Qidong Huang and Jie Huang and Fei Huang and Binyuan Hui and Shutong Jiang and Zhaohai Li and Mingsheng Li and Mei Li and Kaixin Li and Zicheng Lin and Junyang Lin and Xuejing Liu and Jiawei Liu and Chenglong Liu and Yang Liu and Dayiheng Liu and Shixuan Liu and Dunjie Lu and Ruilin Luo and Chenxu Lv and Rui Men and Lingchen Meng and Xuancheng Ren and Xingzhang Ren and Sibo Song and Yuchong Sun and Jun Tang and Jianhong Tu and Jianqiang Wan and Peng Wang and Pengfei Wang and Qiuyue Wang and Yuxuan Wang and Tianbao Xie and Yiheng Xu and Haiyang Xu and Jin Xu and Zhibo Yang and Mingkun Yang and Jianxin Yang and An Yang and Bowen Yu and Fei Zhang and Hang Zhang and Xi Zhang and Bo Zheng and Humen Zhong and Jingren Zhou and Fan Zhou and Jing Zhou and Yuanzhi Zhu and Ke Zhu},
      year={2025},
      eprint={2511.21631},
      archivePrefix={arXiv},
      primaryClass={cs.CV},
      url={https://arxiv.org/abs/2511.21631}, 
}

@misc{sun2026thinkinghurtsmitigatingvisual,
      title={When Thinking Hurts: Mitigating Visual Forgetting in Video Reasoning via Frame Repetition}, 
      author={Xiaokun Sun and Yubo Wang and Haoyu Cao and Linli Xu},
      year={2026},
      eprint={2603.16256},
      archivePrefix={arXiv},
      primaryClass={cs.CV},
      url={https://arxiv.org/abs/2603.16256}, 
}

@inproceedings{sun-etal-2025-mitigating-visual,
    title = "Mitigating Visual Forgetting via Take-along Visual Conditioning for Multi-modal Long {C}o{T} Reasoning",
    author = "Sun, Hai-Long  and
      Sun, Zhun  and
      Peng, Houwen  and
      Ye, Han-Jia",
    editor = "Che, Wanxiang  and
      Nabende, Joyce  and
      Shutova, Ekaterina  and
      Pilehvar, Mohammad Taher",
    booktitle = "Proceedings of the 63rd Annual Meeting of the Association for Computational Linguistics (Volume 1: Long Papers)",
    month = jul,
    year = "2025",
    address = "Vienna, Austria",
    publisher = "Association for Computational Linguistics",
    url = "https://aclanthology.org/2025.acl-long.257/",
    doi = "10.18653/v1/2025.acl-long.257",
    pages = "5158--5171",
    ISBN = "979-8-89176-251-0"
}

@inproceedings{hu-etal-2026-video,
    title = "Video-{MMMU}: Evaluating Knowledge Acquisition from Multidisciplinary Professional Videos",
    author = "Hu, Kairui  and
      Wu, Penghao  and
      Pu, Fanyi  and
      Xiao, Wang  and
      Yue, Xiang  and
      Li, Bo  and
      Zhang, Yuanhan  and
      Liu, Ziwei",
    editor = "Liakata, Maria  and
      Moreira, Viviane P.  and
      Zhang, Jiajun  and
      Jurgens, David",
    booktitle = "Proceedings of the 64th Annual Meeting of the {A}ssociation for {C}omputational {L}inguistics (Volume 1: Long Papers)",
    month = jul,
    year = "2026",
    address = "San Diego, California, United States",
    publisher = "Association for Computational Linguistics",
    url = "https://aclanthology.org/2026.acl-long.1281/",
    doi = "10.18653/v1/2026.acl-long.1281",
    pages = "27798--27828",
    ISBN = "979-8-89176-390-6"
}

@inproceedings{liu-etal-2024-tempcompass,
    title = "{T}emp{C}ompass: Do Video {LLM}s Really Understand Videos?",
    author = "Liu, Yuanxin  and
      Li, Shicheng  and
      Liu, Yi  and
      Wang, Yuxiang  and
      Ren, Shuhuai  and
      Li, Lei  and
      Chen, Sishuo  and
      Sun, Xu  and
      Hou, Lu",
    editor = "Ku, Lun-Wei  and
      Martins, Andre  and
      Srikumar, Vivek",
    booktitle = "Findings of the Association for Computational Linguistics: ACL 2024",
    month = aug,
    year = "2024",
    address = "Bangkok, Thailand",
    publisher = "Association for Computational Linguistics",
    url = "https://aclanthology.org/2024.findings-acl.517/",
    doi = "10.18653/v1/2024.findings-acl.517",
    pages = "8731--8772"
}

@InProceedings{Zhao_2025_CVPR,
    author    = {Zhao, Yilun and Zhang, Haowei and Xie, Lujing and Hu, Tongyan and Gan, Guo and Long, Yitao and Hu, Zhiyuan and Chen, Weiyuan and Li, Chuhan and Xu, Zhijian and Wang, Chengye and Shangguan, Ziyao and Liang, Zhenwen and Liu, Yixin and Zhao, Chen and Cohan, Arman},
    title     = {MMVU: Measuring Expert-Level Multi-Discipline Video Understanding},
    booktitle = {Proceedings of the IEEE/CVF Conference on Computer Vision and Pattern Recognition (CVPR)},
    month     = {June},
    year      = {2025},
    pages     = {8475-8489}
}

@misc{rasheed2025videomathqabenchmarkingmathematicalreasoning,
      title={VideoMathQA: Benchmarking Mathematical Reasoning via Multimodal Understanding in Videos}, 
      author={Hanoona Rasheed and Abdelrahman Shaker and Anqi Tang and Muhammad Maaz and Ming-Hsuan Yang and Salman Khan and Fahad Shahbaz Khan},
      year={2025},
      eprint={2506.05349},
      archivePrefix={arXiv},
      primaryClass={cs.CV},
      url={https://arxiv.org/abs/2506.05349}, 
}

@InProceedings{Feng_2026_CVPR,
    author    = {Feng, Kaituo and Zhang, Manyuan and Li, Hongyu and Fan, Kaixuan and Chen, Shuang and Jiang, Yilei and Zheng, Dian and Sun, Peiwen and Zhang, Yiyuan and Sun, Haoze and Feng, Yan and Pei, Peng and Cai, Xunliang and Yue, Xiangyu},
    title     = {OneThinker: All-in-one Reasoning Model for Image and Video},
    booktitle = {Proceedings of the IEEE/CVF Conference on Computer Vision and Pattern Recognition (CVPR)},
    month     = {June},
    year      = {2026},
    pages     = {5432-5443}
}

@misc{ren2026univrthinkingvisualspace,
      title={UniVR: Thinking in Visual Space for Unified Visual Reasoning}, 
      author={Zhongwei Ren and Yunchao Wei and Yao Zhao and Weibo Gong and Xiao Liu and Anran Wang and Xiangtai Li and Xiaojie Jin},
      year={2026},
      eprint={2607.12800},
      archivePrefix={arXiv},
      primaryClass={cs.CV},
      url={https://arxiv.org/abs/2607.12800}, 
}

@inproceedings{zhang-etal-2025-lmms,
    title = "{LMM}s-Eval: Reality Check on the Evaluation of Large Multimodal Models",
    author = "Zhang, Kaichen  and
      Li, Bo  and
      Zhang, Peiyuan  and
      Pu, Fanyi  and
      Cahyono, Joshua Adrian  and
      Hu, Kairui  and
      Liu, Shuai  and
      Zhang, Yuanhan  and
      Yang, Jingkang  and
      Li, Chunyuan  and
      Liu, Ziwei",
    editor = "Chiruzzo, Luis  and
      Ritter, Alan  and
      Wang, Lu",
    booktitle = "Findings of the Association for Computational Linguistics: NAACL 2025",
    month = apr,
    year = "2025",
    address = "Albuquerque, New Mexico",
    publisher = "Association for Computational Linguistics",
    url = "https://aclanthology.org/2025.findings-naacl.51/",
    doi = "10.18653/v1/2025.findings-naacl.51",
    pages = "881--916",
    ISBN = "979-8-89176-195-7"
}

@InProceedings{Fu_2025_ICCV,
    author    = {Fu, Tianyu and Liu, Tengxuan and Han, Qinghao and Dai, Guohao and Yan, Shengen and Yang, Huazhong and Ning, Xuefei and Wang, Yu},
    title     = {FrameFusion: Combining Similarity and Importance for Video Token Reduction on Large Vision Language Models},
    booktitle = {Proceedings of the IEEE/CVF International Conference on Computer Vision (ICCV)},
    month     = {October},
    year      = {2025},
    pages     = {22654-22663}
}

@InProceedings{Tao_2025_CVPR,
    author    = {Tao, Keda and Qin, Can and You, Haoxuan and Sui, Yang and Wang, Huan},
    title     = {DyCoke: Dynamic Compression of Tokens for Fast Video Large Language Models},
    booktitle = {Proceedings of the IEEE/CVF Conference on Computer Vision and Pattern Recognition (CVPR)},
    month     = {June},
    year      = {2025},
    pages     = {18992-19001}
}

@InProceedings{Ray_2026_CVPR,
    author    = {Ray, Arijit and Abdelkader, Ahmed and Mao, Chengzhi and Plummer, Bryan A. and Saenko, Kate and Krishna, Ranjay and Guibas, Leonidas and Chu, Wen-Sheng},
    title     = {Mull-Tokens:  Modality-Agnostic Latent Thinking},
    booktitle = {Proceedings of the IEEE/CVF Conference on Computer Vision and Pattern Recognition (CVPR) Findings},
    month     = {June},
    year      = {2026},
    pages     = {9477-9488}
}

@misc{chen2025scalingrllongvideos,
      title={Scaling RL to Long Videos}, 
      author={Yukang Chen and Wei Huang and Baifeng Shi and Qinghao Hu and Hanrong Ye and Ligeng Zhu and Zhijian Liu and Pavlo Molchanov and Jan Kautz and Xiaojuan Qi and Sifei Liu and Hongxu Yin and Yao Lu and Song Han},
      year={2025},
      eprint={2507.07966},
      archivePrefix={arXiv},
      primaryClass={cs.CV},
      url={https://arxiv.org/abs/2507.07966}, 
}

@misc{gao2026videosearcherempoweringvideodeep,
      title={VideoSearcher: Empowering Video Deep Research with Multi-Tool Agentic Reasoning via Reinforcement Learning}, 
      author={Zhenkun Gao and Yicheng Bao and Jinlong Peng and Xueheng Li and Theo Huang and Bangwei Liu and Kunquan Li and Zhenye Gan and Tao Hu and Chengjun Xie and Mingqian Yang and Xuanhua He and Zhizhong Zhang and Xin Tan and Chengjie Wang and Yuan Xie},
      year={2026},
      eprint={2607.02927},
      archivePrefix={arXiv},
      primaryClass={cs.CV},
      url={https://arxiv.org/abs/2607.02927}, 
}

@article{shao2024deepseekmath,
  title={Deepseekmath: Pushing the limits of mathematical reasoning in open language models},
  author={Shao, Zhihong and Wang, Peiyi and Zhu, Qihao and Xu, Runxin and Song, Junxiao and Bi, Xiao and Zhang, Haowei and Zhang, Mingchuan and Li, YK and Wu, Yang and others},
  journal={arXiv preprint arXiv:2402.03300},
  year={2024}
}
